\documentclass[11pt,a4paper]{article}
\usepackage[hyperref]{acl2021}
\usepackage{times}
\usepackage{latexsym}

\usepackage{microtype}
\usepackage{url}
\usepackage{algorithm}  
\usepackage{algorithmic} 
\usepackage{graphicx}
\usepackage{enumitem}
\usepackage{amsfonts}
\usepackage{amsmath}
\usepackage{multirow}
\usepackage{array} 
\usepackage{booktabs}  
\usepackage{threeparttable}  
\usepackage{multicol}  
\usepackage{multirow}  
\usepackage{natbib}
\usepackage{epsfig}
\usepackage{algorithm}
\usepackage{algorithmic}
\usepackage{amssymb}
\usepackage[utf8]{inputenc}
\usepackage{cleveref}
\usepackage{verbatim}
\crefname{section}{§}{§§}
\Crefname{section}{§}{§§}

\usepackage{microtype}

\aclfinalcopy 
\def\aclpaperid{1674} 

\title{Preview, Attend and Review: Schema-Aware Curriculum Learning for Multi-Domain Dialog State Tracking}

\author{
Yinpei Dai$^\dag$, Hangyu Li$^\dag$, Yongbin Li$^\dag$\thanks{$^*$Corresponding author}, Jian Sun$^\dag$, Fei Huang$^\dag$, Luo Si$^\dag$, Xiaodan Zhu$^\ddag$\\
    \text{$^\dag$Alibaba Group}\\
    \text{$^\ddag$Ingenuity Labs Research Institute \& ECE, Queen's University}\\
    \texttt{\{yinpei.dyp,hangyu.lhy,shuide.lyb\}@alibaba-inc.com}\\
    \texttt{\{jian.sun,f.huang,luo.si\}@alibaba-inc.com, zhu2048@gmail.com}

}

\date{}

\begin{document}
\maketitle
\begin{abstract}
Existing dialog state tracking (DST) models are trained with dialog data in a random order, neglecting rich structural information in a dataset. In this paper, we propose to use curriculum learning (CL) to better leverage both the curriculum structure and schema structure for  task-oriented dialogs. 
Specifically, we propose a model-agnostic framework called \textbf{S}chema-\textbf{a}ware \textbf{C}urriculum \textbf{L}earning for Dial\textbf{og} State Tracking (SaCLog), which consists of a preview module that pre-trains a DST model with schema information, a curriculum module that optimizes the model with CL, and a review module that augments mispredicted data to reinforce the CL training. 
We show that our proposed approach improves DST performance over both a transformer-based and RNN-based DST model (TripPy and TRADE) and achieves new state-of-the-art results on WOZ2.0 and MultiWOZ2.1. 
\end{abstract}

\section{Introduction}
Dialog state tracking (DST) extracts users' goals in task-oriented dialog systems, where dialog states are often represented in terms of a set of slot-value pairs \cite{williams2014dialog,eric-etal-2020-multiwoz}.
Due to the language variety of multi-turn dialogs, the concepts of slots and values are often indirectly expressed in the conversation (such as co-references, ellipsis, and diverse appearances), which are a major bottleneck for improving DST performance \cite{gao-etal-2019-dialog, hu-etal-2020-sas}. Many existing DST methods have focused on designing better model architectures to tackle the problems \cite{dai2018tracking, wu-etal-2019-transferable,kim-etal-2020-efficient}, but still
neglect the full exploitation of two important aspects of structural information.  

\begin{figure}[t]
    \centering
    \includegraphics[width=0.46\textwidth]{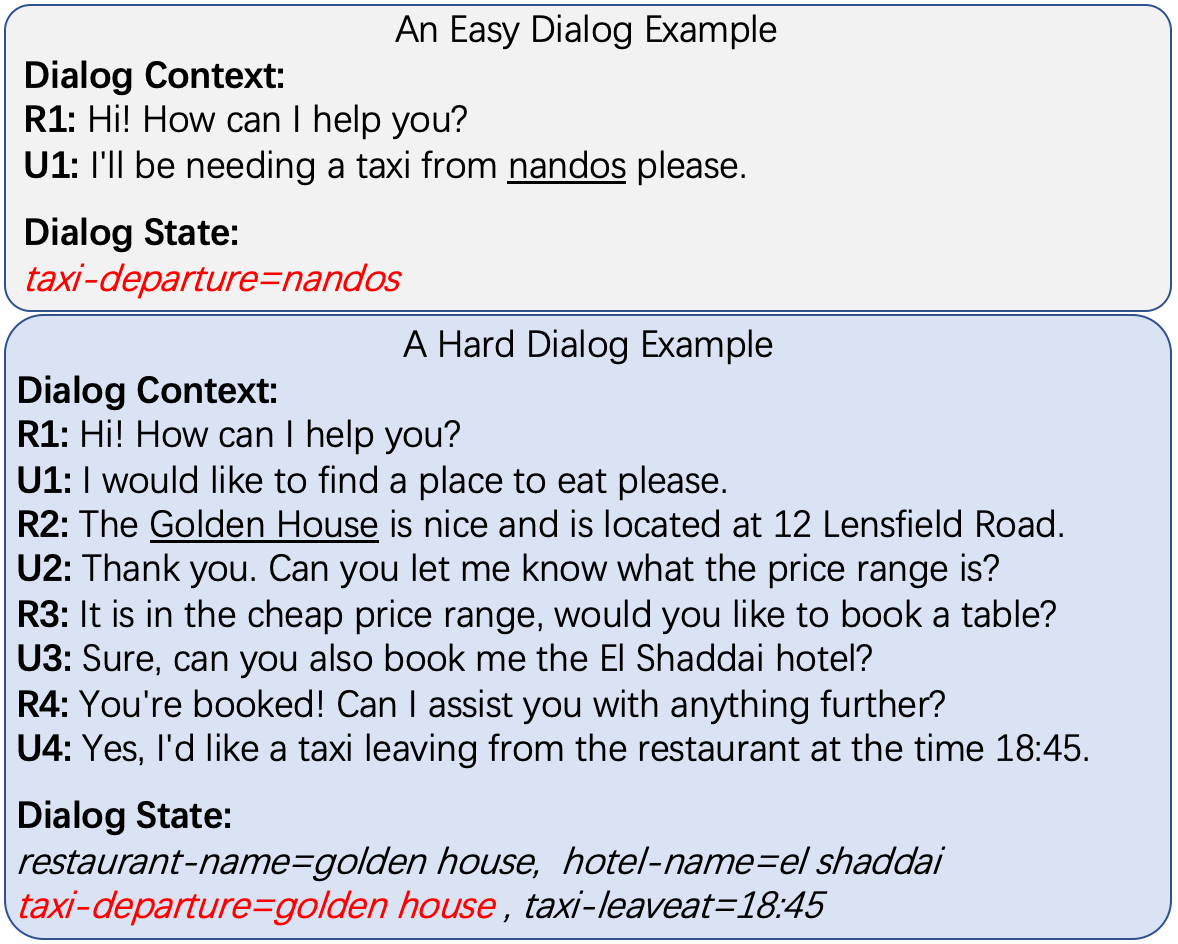}
    \vspace{-0.1cm}
    \caption{An easy and a hard dialog example for DST.}
    \label{easy_hard_examples}
    \vspace{-0.6cm}
\end{figure}

The first is \textit{curriculum  structure} in a dataset. 
Such a structure relies on a measure of the difficulty of examples, which can be used to guide the model training in an easy-to-hard manner, imitating the meaningful learning order in human curricula. 
This paradigm is called curriculum learning (CL) \cite{bengio2009curriculum} and has been shown useful in various other problems \cite{wang2020comprehensive}. DST training examples also vary greatly in their difficulty levels. As shown in Figure \ref{easy_hard_examples}, for the same slot \textit{`taxi-departure'}, a user can either inform its value \textit{`nandos'} explicitly in a simple utterance or convey her intention implicitly via multi-round interactions, requiring a complex inference process to find the value \textit{`golden house'} referred by the slot \textit{`restaurant-name'}. However, CL has been rarely studied in DST, and models are often trained with dialog data in a random order.


In addition, \textit{schema structure} is prominent in multi-domain task-oriented dialogs. A schema is specified by a collection of all possible slots and their values, which describes semantic relations among them. Some previous work utilized the structure via an extra schema graph in a regular  training process \cite{chen2020schema,zhu-etal-2020-efficient, wu-etal-2020-gcdst}. 
We propose to incorporate schema information into CL through a pre-curriculum process, in which a DST model can be pre-trained with schema-related objectives to prepare for upcoming DST examples. To reinforce the CL training, we can also expand those examples with frequent mispredictions during CL based upon the schema, enabling the model to accumulate more experience and perform better on similar cases. 

Built on these motivations, we propose a novel framework named as \textbf{S}chema-\textbf{a}ware \textbf{C}urriculum \textbf{L}earning for Dial\textbf{og} State Tracking (SaCLog), which consists of three components: 1) a \textbf{preview module} that pre-trains the base part of a DST model (e.g., BERT and RNN) with objectives capturing the connections between the schema and dialog contexts,  2) a \textbf{curriculum module} that organizes training data from easy to hard and optimizes the model with CL, and 3) a \textbf{review module} which leverages schema-based data augmentation to extend mispredicted data to boost the CL training process further. The proposed approach is model-agnostic, in the sense that it can be incorporated into different DST models.
To the best of our knowledge, this is the first attempt to apply CL to the DST task. 
We show that our proposed approach improves DST performance over both a transformer-based and RNN-based DST model (TripPy and TRADE) and achieves new state-of-the-art results on WOZ2.0 and MultiWOZ2.1. 

\section{Problem Formulation} 
We denote a dialog context containing $t$ turns as $X_t=\{(R_1, U_1),...,(R_t, U_t)\}$, where $R_i$ and $U_i$ represent system and user utterance at the $i$-th turn respectively. DST is tasked to extract turn-level or discourse-level dialog states in the form of a set of slot-value pairs given $X_t$. A turn-level dialog state $Y_t=\{(s, v_t), s\in \mathcal{S}\}$ is the slot-value pairs extracted only from $(R_t, U_t)$ at current turn $t$, where $\mathcal{S}$ is a predefined  set of slot $s$ in the schema
and $v_t$ is the corresponding value\footnote{Each $s$ contains two special values, \textit{none} and \textit{dontcare}, indicating $s$ has no values and can take any values respectively.} of the slot $s$. A discourse-level dialog state $Z_t$ is the accumulation of $L_t$, representing all slot-value pairs that have been expressed over the course of the dialog until the $t$-th turn. We denote a dialog data for DST as $d_t=\{X_t,Y_t,Z_t\}$ and the training dataset as $\mathcal{D}$.

\section{Schema-Aware Curriculum Learning}

\begin{figure*}[t]
    \centering
    \includegraphics[width=0.99\textwidth]{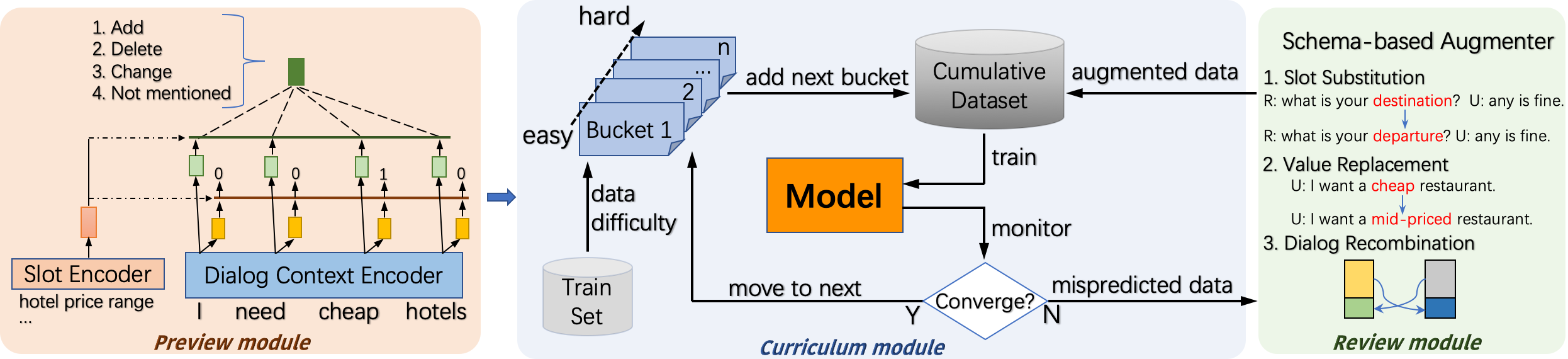}
    \caption{An overview of the SaCLog training procedures.}
    \label{fig:SACL}
    \vspace{-0.4cm}
\end{figure*}


In this section, we first introduce the core curriculum module about how to apply the basic CL to the DST task; we then describe the preview and review module, which exploit the schema structure to facilitate the CL training process.
The overall framework of SaCLog is shown in Figure \ref{fig:SACL}.

\vspace{-0.1cm}
\subsection{Curriculum Learning for DST}
\vspace{-0.05cm}

We propose curriculum learning for DST and design two sub-modules: a \textbf{difficulty scorer} that measures the difficulty level of a dialog example with respect to a DST model, as well as  a \textbf{training scheduler} module that arranges the scored data as a sequence of easy-to-hard training stages. 

\subsubsection{The Difficulty Scorer}
As a dialog example could be intuitively \textit{complex} for humans  or inherently \textit{difficult} for neural networks (NNs), both model-based and rule-based scores should be considered. We propose to use a hybrid scoring function that combines the advantages of model predictions and rules. 

For model-based difficulty, we predict scores in a cross-validation-like manner. We divide $\mathcal{D}$ into $K$ equal-sized subsets, where $K-1$ subsets are used to train a DST model to predict the remaining one. This process is repeated $K$ times until every subset is predicted. 
The score $r^{mod}_t \in [0,1] $ is computed based on the average accuracy of all mentioned slots (whose values are not \textit{none}) in $Y_t$ for each $d_t$. In our experiment, we train six models with the same architecture and different initialization seeds to obtain the mean value $\overline{r}^{mod}_t$ of model scores.

For rule-based difficulty, we consider 4 factors to fuse human prior knowledge about DST into our curriculum design: 1) current dialog turn number $t$;  2) the total token number of $(R_t, U_t)$; 3) the number of mentioned name entities like `hotel names' in $Z_t$; 4) the number of newly added or changed slots in $Y_t$. We set the maximum values of above factors as 7/50/4/6 respectively, and normalize all factors into $r_t^{rul,i}\in [0,1]$, where $i$ indicates the $i$-th factor. 

Finally, the hybrid difficulty score is calculated jointly as $r^{hyb}_t = \alpha_0 \overline{r}^{mod}_t + \sum_{i=1}^4 \alpha_i r^{rul,i}_t$, where $r^{hyb}\in[0,1]$ and  $\sum_{i=0}^4\alpha_i=1$. 

\subsubsection{The Training Scheduler}
We adopt a widely used strategy called \textit{baby step} \cite{spitkovsky-etal-2010-baby} to organize the scored data for CL. Specifically, we divide the score uniformly into $N$ intervals  and distribute the sorted data into $N$ buckets accordingly. The optimization starts from the easiest bucket as the initial training stage. After reaching a fixed number of maximum epochs or convergence, the next bucket is merged into the current training subset and shuffled for the next training stage. 
In our experiment, we set the maximum number of epochs as 3, and treat as the convergence if the training loss ceases to decrease and the loss value is within a threshold 15 for 100 steps.
As the subset accumulates until all buckets are aggregated, we then continue to train the model for several extra epochs. 



\subsection{The Preview Module} 
\vspace{-0.05cm}
In human learning, previewing learning materials helps develop an overall picture of what will be covering and can bring benefits to the learning process. In our task here, we propose new pre-training objectives to learn structural inductive bias of the schema structure. Specifically, our \textit{preview} module contains a slot encoder to compute a slot embedding $e_{s}$ for each input slot $s$, and a dialog context encoder to extract the hidden states of $X_t$ as $E_{t}=[e^1_t,e^2_t,...]$, then we have:

\vspace{-0.5cm}
\begin{equation}
\begin{aligned}
B^s_t & = [\phi_1^{sig}(e_s \oplus e^1_t), \phi_1^{sig}(e_s \oplus e^2_t),...] \\
c^s_t & = \phi_4^{sft}\left(e_s \oplus \text{Att}(e_s, E_t)\right)\\
\end{aligned}
\end{equation}
\vspace{-0.4cm}

\noindent where $\text{Att}(k, V)$ is the attention function using the vector $k$ to query the vector sequence $V$ to get a context vector and $\oplus$ the vector concatenation. $\phi_{d}^{sig}(\cdot)$ and $\phi_{d}^{sft}(\cdot)$ 
denote an FNN with one hidden layer having the same size as input layer, where the output layer is of size $d$, and is sigmoid and softmax respectively.
$B^s_t$ is a binary sequence indicates which span of $X_t$ belongs to the value of $s$, while $c^s_t$ is the classification logits indicates whether $s$ is added, deleted, changed, or not mentioned in $Y_t$.

Therefore, for each slot $s$, we have a binary sequence loss $L_{seq}$ and a classification loss $L_{cls}$ to optimize. 
Such pre-training objectives help the encoders understand how a slot is roughly operated in the current dialog context and connected with all possible tokens regarding its values in the schema.  
The dialog context encoder is used for the parameter initialization of the base part of a DST model.
The pre-trained corpus is constructed from MultiWOZ2.1 dialogs  \cite{eric-etal-2020-multiwoz} and the off-the-shelf synthesized dialogs  \cite{campagna-etal-2020-zero}, which contains 337,346 dialog data in total. 

We also leverage the language modelling (LM) loss as an auxiliary loss $L_{aux}$ to learn contextual representations of natural language. To be specific, we use the MLM loss \cite{devlin-etal-2019-bert} as $L_{aux}$ for transformer-based DST modes and the summation of both forward and backward LM losses \cite{peters-etal-2018-deep} for RNN-based DST models. We only use the original MultiWOZ2.1 dialogs to optimize $L_{aux}$, considering that synthesized data is not suitable for natural language modelling. However, both the original and synthesized data are used to optimize $L_{seq}$ and $L_{cls}$. 

\vspace{-0.03cm}
\subsection{The Review Module}
\vspace{-0.03cm}
The process of review often help a learner consolidate difficult concepts newly learned. We design a \textit{review} module to consider mispredicted examples as the concepts that the DST model has not grasped during CL, and utilize a schema-based data augmenter to produce similar cases from the examples. Specifically, the DST model is monitored at each stage of the CL training process. If a model is not converged at the end of an epoch in a training stage, we choose the top 10\% incorrectly predicted examples according to their training losses as the resource to enlarge the cumulative dataset.
The schema-based data augmenter uses three   practical techniques to generate data as follows:
\paragraph{Slot Substitution.} A mentioned slot name in $(R_t,U_t)$ is changed into another slot name when its value is \textit{dontcare}. Specifically, we first collect a word set for each slot name,  e.g. \textit{\{`arrive', `arriving', `arrived'\}} for the slot \textit{`taxi-arriveby'}. Then, for a dialog data $d_t$ where $Y_t$ contains a slot $s$ with the value \textit{dontcare}, we substitute the word of $s$ in the utterance with some word of another slot $s'$ that is of the same domain and not mentioned in $Y_t$.

\paragraph{Value Replacement.} A slot's value is replaced with another proper one when the value is explicitly contained in $U_t$. Specifically, we leverage the predefined schema in the dialog dataset to produce a value set for each slot and use the label map in \cite{heck-etal-2020-trippy} to figure out the position of value span within the utterance. The target value is then replaced  with another one of the same slot. 


\paragraph{Dialog Recombination.} To recombine the dialog data $d_t$, we randomly search another dialog data in $D$ that possesses the same mentioned slots (whose values are not \textit{none}) in $Y_t$. We then cut and stitch their history $X_{t-1}$ and current utterances $(R_t,U_t)$, and exchange their $Y_t$ to produce two new dialog data.


\vspace{-0.1cm}
\section{Experiments}
\vspace{-0.1cm}
\label{exp}
Two popular datasets, WOZ2.0 \cite{wen-etal-2017-network} and MultiWOZ2.1 \cite{eric-etal-2020-multiwoz}, are used to verify our approach. WOZ2.0 is a single-domain dataset with 1,200 dialogs and 3 slots. MultiWOZ2.1 is a multi-domain dialog dataset with 10,438 dialogs, where there are 30 slots spanning 7 domains. 
The data splits (train/valid/test) of WOZ2.0 and MultiWOZ2.1 are 600/200/400 and 8438/1000/1000, respectively. 
We use the joint goal accuracy (JGA), the ratio of dialog data whose $Z_t$ is correct, as the evaluation metric. 
We apply SaCLog onto TripPy \cite{heck-etal-2020-trippy}, a transformer-based DST model, and TRADE \cite{wu-etal-2019-transferable}, an RNN-based DST model, to show its effect. 
The slot encoder and the dialog context encoder are weight-shared.
We use a BERT$_{base}$ as the encoder and the \texttt{[CLS]} embedding as the slot embedding in TripPy, and use a bi-GRU as the encoder and the concatenation of the first and last hidden state as the slot embedding for TRADE. We also follow TripPy to add 2 new slot operations (i.e. \textit{refer/dontcare}) into the classification types of $L_{cls}$.
\paragraph{Implementation Details.} For the preview module, we use Adam \cite{kingma2014adam} with a fixed learning rate 3e-5 for 3 epochs in the pre-training. The batch size for $L_{aux}$ is 14 and the batch size for $L_{seq}$ and $L_{cls}$ is 64.  For the curriculum module, we perform a warm-up strategy for Adam optimizer with a maximum learning rate 1e-4. Before CL, we train models on full dataset for 2 epochs.
After all subsets are accumulated, we then train for 10 extra epochs with a minimum learning rate 1e-6. We set the bucket number $N=10$ and the crossed fold $K=5$. The batch size is 36 and the maximum length is 256. To simplify the review process, we conduct data augmentation after the CL training is finished.


\subsection{Performance of TripPy+SaCLog}
\begin{table}[t]
    \centering
    \scalebox{0.75}{
    \begin{tabular}{l|c|c}
        \toprule
         Models & MultiWOZ2.1 & WOZ2.0\\
         \midrule
         GLAD \small\cite{zhong-etal-2018-global} & 35.57\%$^{**}$ & 88.1$\pm$0.4\% \\ 
         SUMBT \small\cite{lee-etal-2019-sumbt} & 46.65\%$^{**}$ & 91.0$\pm$1.0\% \\
         DST-picklist \small\cite{zhang2019find} & 53.30\% & -- \\
         Trippy \small\cite{heck-etal-2020-trippy}  & 55.29$\pm$0.28\% & 92.7$\pm$0.2\% \\
         SimpleTOD\scriptsize\cite{hosseini2020simple}  & 55.72\% & -- \\
         CHAN \small\cite{shan-etal-2020-contextual}  & 58.55\% & -- \\
         TripPy + ConvBERT & 58.70\% & 93.1$\pm$0.3\%$^*$\\ 
         TripPy + CoCoAug & 60.53\% & -- \\
         \hline
         TripPy + SaCLog & \textbf{60.61$\pm$0.31\%} & \textbf{94.2$\pm$0.2\%} \\
         \bottomrule
    \end{tabular}}
    \vspace{-0.15cm}
    \caption{DST Results on MultiWOZ2.1 and WOZ2.0 in JGA. $*$ Our implemetation. $**$ MultiWOZ2.0 results.}
    \label{tab:all_results}
    \vspace{-0.1cm}
\end{table}
\begin{table}[t]
    \centering
    \scalebox{0.8}{
    \begin{tabular}{l|c}
        \toprule
         Models & JGA \\
         \midrule
         TripPy (ours) & 58.17$\pm$0.25\% \\
         \ \ \ \  + CL \small (rule-based) & 58.38$\pm$0.17\% \\
         \ \ \ \  + CL \small (model-based) & 58.71$\pm$0.21\% \\
         \ \ \ \  + CL \small (hybrid) & \textbf{58.85$\pm$0.23\%} \\
         \hline 
         \ \ \ \  + SaCLog \small (w/o. review) & 60.19$\pm$0.26\% \\
         \ \ \ \  + SaCLog \small (w/o. preview) & 60.23$\pm$0.34\% \\
         \ \ \ \  + SaCLog & \textbf{60.61$\pm$0.31\%} \\
         \bottomrule
    \end{tabular}}
    \vspace{-0.15cm}
    \caption{Ablation results on MultiWOZ2.1. +CL means adding the curriculum module only.}
    \label{tab:multiwoz_ablation}
    \vspace{-0.4cm}
\end{table}

Tables 1 shows the results of our
approach comparing to various baselines. Based upon TripPy, we obtain state-of-the-art performance on both datasets with SaCLog. The two closest baselines\footnote{We implemented SaCLog upon these two methods, but no significant gains are observed. We conjecture that this is due to SaCLog has already largely exploited TripPy's potential so that the additional improvement of the two methods is limited.}, ConvBERT \cite{mehri2020dialoglue} and CoCoAug \cite{li2020coco}, are also built upon TripPy, where ConvBERT enhances its performance by using external large-scale conversational corpora to pre-train a BERT$_{base}$ and CoCoAug leverages a delicate counter-factual augmentation skill to produce much larger training data. Our method, however, benefits from the CL framework and improves TripPy by utilizing the preview and review modules.

\paragraph{Ablation Study}
To examine how SaCLog facilitates DST training, we conduct detailed ablation experiments on MultiWOZ2.1, as shown in Table 2. In our re-implementation, we improve the basic TripPy by around 3\% JGA via  training for longer epochs (30 vs.10) and pre-training a BERT$_{base}$ on MultiWOZ2.1 corpus only with the MLM loss. First, we investigate the influence of difficulty scores by adding the curriculum module and utilizing the same pre-trained BERT$_{base}$. As we can see, using the hybrid difficulty score achieves better JGA (58.85\%) than using either single score, indicating that both model prediction and human knowledge are necessary.
When incorporating the other two modules in the CL framework, the performance is greatly boosted further. The combination of both modules increases the JGA by 1.76\%, suggesting that the schema-aware pre-training and dialog augmentation are crucial for improving DST performance in the CL training.


\subsection{Performance of TRADE+SaCLog}
We also apply SaCLog to the classical RNN-based generative DST model, TRADE. As Table 3 shows, SaCLog improves TRADE by around 3$\sim$4\% JGA on both datasets, demonstrating the effectiveness of SaCLog on different types of base DST models.

\begin{table}[t]
    \centering
    \scalebox{0.8}{
    \begin{tabular}{l|c|c}
        \toprule
         Model & MultiWOZ2.1 & WOZ2.0 \\
         \midrule
         TRADE & 45.6\%$^*$ & 88.3$\pm$0.6\% \\ 
         \ \ \ \ + SaCLog & \textbf{49.3$\pm$0.5\%} & \textbf{91.1$\pm$0.4\%} \\
         \bottomrule
    \end{tabular}}
    \vspace{-0.15cm}
    \caption{Results of TRADE+SaCLog on MultiWOZ2.1 and WOZ2.0. * Reported in \cite{eric-etal-2020-multiwoz}.}
    \label{tab:trade_results}
    \vspace{-0.4cm}
\end{table}

\section{Related Work}
\vspace{-0.05cm}
Curriculum Learning (CL) has attracted increasing research interests in various NLP tasks, such as  machine translation \cite{liu-etal-2020-norm, zhou-etal-2020-uncertainty}, 
general language understanding \cite{xu-etal-2020-curriculum}, 
reading comprehension \cite{tay-etal-2019-simple} and open-domain chatbots \cite{bao2020plato,cai2020learning, su2020dialogue}. Yet, the research on using CL in task-oriented dialog systems is limited. There has been some work \cite{saito-2018-curriculum, zhao2020automatic} on using CL in dialog policy learning, but applying CL to DST has not been investigated.

Learning a structural inductive bias
during pre-training has been shown beneficial in downstream tasks that require parsing semantics, such as text-to-SQL \cite{yu2020grappa} and table cell recognition \cite{wang2020structure}. There are also many works \cite{hou-etal-2018-sequence, yoo-etal-2020-variational, yin2020dialog} on dialog augmentation. We aim to integrate these 
methods
to build a general CL framework for DST. 

\vspace{-0.05cm}
\section{Conclusion}
\vspace{-0.05cm}
In this paper, we propose a model-agnostic framework named as schema-aware curriculum learning for DST, which exploits both the curriculum structure and the  schema structure in task-oriented dialogs and shows to substantially improve DST performances. In the future, we plan to investigate CL approaches on other dialog modeling tasks.

\vspace{-0.1cm}
\section*{Acknowledgments}
\vspace{-0.1cm}
The research of the last author is supported by the Natural Sciences and Engineering Research Council of Canada (NSERC).

\bibliographystyle{acl_natbib}
\bibliography{anthology,acl2021}

\begin{thebibliography}{40}
\expandafter\ifx\csname natexlab\endcsname\relax\def\natexlab#1{#1}\fi

\bibitem[{Bao et~al.(2020)Bao, He, Wang, Wu, Wang, Wu, Guo, Liu, and
  Xu}]{bao2020plato}
Siqi Bao, Huang He, Fan Wang, Hua Wu, Haifeng Wang, Wenquan Wu, Zhen Guo,
  Zhibin Liu, and Xinchao Xu. 2020.
\newblock \href {https://arxiv.org/pdf/2006.16779} {Plato-2: Towards building
  an open-domain chatbot via curriculum learning}.
\newblock \emph{arXiv preprint arXiv:2006.16779}.

\bibitem[{Bengio et~al.(2009)Bengio, Louradour, Collobert, and
  Weston}]{bengio2009curriculum}
Yoshua Bengio, J{\'e}r{\^o}me Louradour, Ronan Collobert, and Jason Weston.
  2009.
\newblock \href {https://dl.acm.org/doi/abs/10.1145/1553374.1553380}
  {Curriculum learning}.
\newblock In \emph{Proceedings of the 26th annual international conference on
  machine learning}, pages 41--48.

\bibitem[{Cai et~al.(2020)Cai, Chen, Zhang, Song, Zhao, Li, Duan, and
  Yin}]{cai2020learning}
Hengyi Cai, Hongshen Chen, Cheng Zhang, Yonghao Song, Xiaofang Zhao, Yangxi Li,
  Dongsheng Duan, and Dawei Yin. 2020.
\newblock \href {https://arxiv.org/abs/2003.00639} {Learning from easy to
  complex: Adaptive multi-curricula learning for neural dialogue generation}.
\newblock In \emph{Proceedings of the AAAI Conference on Artificial
  Intelligence}, volume~34, pages 7472--7479.

\bibitem[{Campagna et~al.(2020)Campagna, Foryciarz, Moradshahi, and
  Lam}]{campagna-etal-2020-zero}
Giovanni Campagna, Agata Foryciarz, Mehrad Moradshahi, and Monica Lam. 2020.
\newblock \href {https://doi.org/10.18653/v1/2020.acl-main.12} {Zero-shot
  transfer learning with synthesized data for multi-domain dialogue state
  tracking}.
\newblock In \emph{Proceedings of the 58th Annual Meeting of the Association
  for Computational Linguistics}, pages 122--132, Online. Association for
  Computational Linguistics.

\bibitem[{Chen et~al.(2020)Chen, Lv, Wang, Zhu, Tan, and Yu}]{chen2020schema}
Lu~Chen, Boer Lv, Chi Wang, Su~Zhu, Bowen Tan, and Kai Yu. 2020.
\newblock \href {https://aaai.org/ojs/index.php/AAAI/article/view/6250/6106}
  {Schema-guided multi-domain dialogue state tracking with graph attention
  neural networks}.
\newblock In \emph{Proceedings of the AAAI Conference on Artificial
  Intelligence}, volume~34, pages 7521--7528.

\bibitem[{Dai et~al.(2018)Dai, Ou, Ren, and Yu}]{dai2018tracking}
Yinpei Dai, Zhijian Ou, Dawei Ren, and Pengfei Yu. 2018.
\newblock \href {https://arxiv.org/abs/1711.03381} {Tracking of enriched dialog
  states for flexible conversational information access}.
\newblock In \emph{2018 IEEE International Conference on Acoustics, Speech and
  Signal Processing (ICASSP)}, pages 6139--6143. IEEE.

\bibitem[{Devlin et~al.(2019)Devlin, Chang, Lee, and
  Toutanova}]{devlin-etal-2019-bert}
Jacob Devlin, Ming-Wei Chang, Kenton Lee, and Kristina Toutanova. 2019.
\newblock \href {https://doi.org/10.18653/v1/N19-1423} {{BERT}: Pre-training of
  deep bidirectional transformers for language understanding}.
\newblock In \emph{Proceedings of the 2019 Conference of the North {A}merican
  Chapter of the Association for Computational Linguistics: Human Language
  Technologies, Volume 1 (Long and Short Papers)}, pages 4171--4186,
  Minneapolis, Minnesota. Association for Computational Linguistics.

\bibitem[{Eric et~al.(2020)Eric, Goel, Paul, Sethi, Agarwal, Gao, Kumar, Goyal,
  Ku, and Hakkani-Tur}]{eric-etal-2020-multiwoz}
Mihail Eric, Rahul Goel, Shachi Paul, Abhishek Sethi, Sanchit Agarwal, Shuyang
  Gao, Adarsh Kumar, Anuj Goyal, Peter Ku, and Dilek Hakkani-Tur. 2020.
\newblock \href {https://www.aclweb.org/anthology/2020.lrec-1.53} {{M}ulti{WOZ}
  2.1: A consolidated multi-domain dialogue dataset with state corrections and
  state tracking baselines}.
\newblock In \emph{Proceedings of the 12th Language Resources and Evaluation
  Conference}, pages 422--428, Marseille, France. European Language Resources
  Association.

\bibitem[{Gao et~al.(2019)Gao, Sethi, Agarwal, Chung, and
  Hakkani-Tur}]{gao-etal-2019-dialog}
Shuyang Gao, Abhishek Sethi, Sanchit Agarwal, Tagyoung Chung, and Dilek
  Hakkani-Tur. 2019.
\newblock \href {https://doi.org/10.18653/v1/W19-5932} {Dialog state tracking:
  A neural reading comprehension approach}.
\newblock In \emph{Proceedings of the 20th Annual SIGdial Meeting on Discourse
  and Dialogue}, pages 264--273, Stockholm, Sweden. Association for
  Computational Linguistics.

\bibitem[{Heck et~al.(2020)Heck, van Niekerk, Lubis, Geishauser, Lin, Moresi,
  and Gasic}]{heck-etal-2020-trippy}
Michael Heck, Carel van Niekerk, Nurul Lubis, Christian Geishauser, Hsien-Chin
  Lin, Marco Moresi, and Milica Gasic. 2020.
\newblock \href {https://www.aclweb.org/anthology/2020.sigdial-1.4}
  {{T}rip{P}y: A triple copy strategy for value independent neural dialog state
  tracking}.
\newblock In \emph{Proceedings of the 21th Annual Meeting of the Special
  Interest Group on Discourse and Dialogue}, pages 35--44, 1st virtual meeting.
  Association for Computational Linguistics.

\bibitem[{Hosseini-Asl et~al.(2020)Hosseini-Asl, McCann, Wu, Yavuz, and
  Socher}]{hosseini2020simple}
Ehsan Hosseini-Asl, Bryan McCann, Chien-Sheng Wu, Semih Yavuz, and Richard
  Socher. 2020.
\newblock \href {https://openreview.net/forum?id=wb114Jun30a} {A simple
  language model for task-oriented dialogue}.
\newblock \emph{Thirty-forth Conference on Neural Information Processing
  Systems}.

\bibitem[{Hou et~al.(2018)Hou, Liu, Che, and Liu}]{hou-etal-2018-sequence}
Yutai Hou, Yijia Liu, Wanxiang Che, and Ting Liu. 2018.
\newblock \href {https://www.aclweb.org/anthology/C18-1105}
  {Sequence-to-sequence data augmentation for dialogue language understanding}.
\newblock In \emph{Proceedings of the 27th International Conference on
  Computational Linguistics}, pages 1234--1245, Santa Fe, New Mexico, USA.
  Association for Computational Linguistics.

\bibitem[{Hu et~al.(2020)Hu, Yang, Chen, He, and Yu}]{hu-etal-2020-sas}
Jiaying Hu, Yan Yang, Chencai Chen, Liang He, and Zhou Yu. 2020.
\newblock \href {https://doi.org/10.18653/v1/2020.acl-main.567} {{SAS}:
  Dialogue state tracking via slot attention and slot information sharing}.
\newblock In \emph{Proceedings of the 58th Annual Meeting of the Association
  for Computational Linguistics}, pages 6366--6375, Online. Association for
  Computational Linguistics.

\bibitem[{Kim et~al.(2020)Kim, Yang, Kim, and Lee}]{kim-etal-2020-efficient}
Sungdong Kim, Sohee Yang, Gyuwan Kim, and Sang-Woo Lee. 2020.
\newblock \href {https://doi.org/10.18653/v1/2020.acl-main.53} {Efficient
  dialogue state tracking by selectively overwriting memory}.
\newblock In \emph{Proceedings of the 58th Annual Meeting of the Association
  for Computational Linguistics}, pages 567--582, Online. Association for
  Computational Linguistics.

\bibitem[{Kingma and Ba(2015)}]{kingma2014adam}
Diederik~P Kingma and Jimmy Ba. 2015.
\newblock \href {https://arxiv.org/abs/1412.6980} {Adam: A method for
  stochastic optimization}.
\newblock \emph{International Conference on Learning Representations}.

\bibitem[{Lee et~al.(2019)Lee, Lee, and Kim}]{lee-etal-2019-sumbt}
Hwaran Lee, Jinsik Lee, and Tae-Yoon Kim. 2019.
\newblock \href {https://doi.org/10.18653/v1/P19-1546} {{SUMBT}: Slot-utterance
  matching for universal and scalable belief tracking}.
\newblock In \emph{Proceedings of the 57th Annual Meeting of the Association
  for Computational Linguistics}, pages 5478--5483, Florence, Italy.
  Association for Computational Linguistics.

\bibitem[{Li et~al.(2021)Li, Yavuz, Hashimoto, Li, Niu, Rajani, Yan, Zhou, and
  Xiong}]{li2020coco}
Shiyang Li, Semih Yavuz, Kazuma Hashimoto, Jia Li, Tong Niu, Nazneen Rajani,
  Xifeng Yan, Yingbo Zhou, and Caiming Xiong. 2021.
\newblock \href {https://openreview.net/forum?id=eom0IUrF__F} {Coco:
  Controllable counterfactuals for evaluating dialogue state trackers}.
\newblock \emph{International Conference on Learning Representations}.

\bibitem[{Liu et~al.(2020)Liu, Lai, Wong, and Chao}]{liu-etal-2020-norm}
Xuebo Liu, Houtim Lai, Derek~F. Wong, and Lidia~S. Chao. 2020.
\newblock \href {https://doi.org/10.18653/v1/2020.acl-main.41} {Norm-based
  curriculum learning for neural machine translation}.
\newblock In \emph{Proceedings of the 58th Annual Meeting of the Association
  for Computational Linguistics}, pages 427--436, Online. Association for
  Computational Linguistics.

\bibitem[{Mehri et~al.(2020)Mehri, Eric, and Hakkani-Tur}]{mehri2020dialoglue}
Shikib Mehri, Mihail Eric, and Dilek Hakkani-Tur. 2020.
\newblock \href {https://arxiv.org/abs/2009.13570} {Dialoglue: A natural
  language understanding benchmark for task-oriented dialogue}.
\newblock \emph{arXiv preprint arXiv:2009.13570}.

\bibitem[{Peters et~al.(2018)Peters, Neumann, Iyyer, Gardner, Clark, Lee, and
  Zettlemoyer}]{peters-etal-2018-deep}
Matthew Peters, Mark Neumann, Mohit Iyyer, Matt Gardner, Christopher Clark,
  Kenton Lee, and Luke Zettlemoyer. 2018.
\newblock \href {https://doi.org/10.18653/v1/N18-1202} {Deep contextualized
  word representations}.
\newblock In \emph{Proceedings of the 2018 Conference of the North {A}merican
  Chapter of the Association for Computational Linguistics: Human Language
  Technologies, Volume 1 (Long Papers)}, pages 2227--2237, New Orleans,
  Louisiana. Association for Computational Linguistics.

\bibitem[{Saito(2018)}]{saito-2018-curriculum}
Atsushi Saito. 2018.
\newblock \href {https://doi.org/10.18653/v1/W18-5707} {Curriculum learning
  based on reward sparseness for deep reinforcement learning of task completion
  dialogue management}.
\newblock In \emph{Proceedings of the 2018 {EMNLP} Workshop {SCAI}: The 2nd
  International Workshop on Search-Oriented Conversational {AI}}, pages 46--51,
  Brussels, Belgium. Association for Computational Linguistics.

\bibitem[{Shan et~al.(2020)Shan, Li, Zhang, Meng, Feng, Niu, and
  Zhou}]{shan-etal-2020-contextual}
Yong Shan, Zekang Li, Jinchao Zhang, Fandong Meng, Yang Feng, Cheng Niu, and
  Jie Zhou. 2020.
\newblock \href {https://doi.org/10.18653/v1/2020.acl-main.563} {A contextual
  hierarchical attention network with adaptive objective for dialogue state
  tracking}.
\newblock In \emph{Proceedings of the 58th Annual Meeting of the Association
  for Computational Linguistics}, pages 6322--6333, Online. Association for
  Computational Linguistics.

\bibitem[{Spitkovsky et~al.(2010)Spitkovsky, Alshawi, and
  Jurafsky}]{spitkovsky-etal-2010-baby}
Valentin~I. Spitkovsky, Hiyan Alshawi, and Daniel Jurafsky. 2010.
\newblock \href {https://www.aclweb.org/anthology/N10-1116} {From baby steps to
  leapfrog: How {``}less is more{''} in unsupervised dependency parsing}.
\newblock In \emph{Human Language Technologies: The 2010 Annual Conference of
  the North {A}merican Chapter of the Association for Computational
  Linguistics}, pages 751--759, Los Angeles, California. Association for
  Computational Linguistics.

\bibitem[{Su et~al.(2020)Su, Cai, Zhou, Lin, Baker, Cao, Shi, Collier, and
  Wang}]{su2020dialogue}
Yixuan Su, Deng Cai, Qingyu Zhou, Zibo Lin, Simon Baker, Yunbo Cao, Shuming
  Shi, Nigel Collier, and Yan Wang. 2020.
\newblock \href {https://arxiv.org/abs/2012.14756} {Dialogue response selection
  with hierarchical curriculum learning}.
\newblock \emph{59th Annual Meeting of the Association for Computational
  Linguistics}.

\bibitem[{Tay et~al.(2019)Tay, Wang, Luu, Fu, Phan, Yuan, Rao, Hui, and
  Zhang}]{tay-etal-2019-simple}
Yi~Tay, Shuohang Wang, Anh~Tuan Luu, Jie Fu, Minh~C. Phan, Xingdi Yuan, Jinfeng
  Rao, Siu~Cheung Hui, and Aston Zhang. 2019.
\newblock \href {https://doi.org/10.18653/v1/P19-1486} {Simple and effective
  curriculum pointer-generator networks for reading comprehension over long
  narratives}.
\newblock In \emph{Proceedings of the 57th Annual Meeting of the Association
  for Computational Linguistics}, pages 4922--4931, Florence, Italy.
  Association for Computational Linguistics.

\bibitem[{Wang et~al.(2021)Wang, Chen, and Zhu}]{wang2020comprehensive}
Xin Wang, Yudong Chen, and Wenwu Zhu. 2021.
\newblock \href {https://arxiv.org/abs/2010.13166} {A survey on curriculum
  learning}.
\newblock \emph{IEEE Transactions on Pattern Analysis and Machine Intelligence
  (TPAMI)}.

\bibitem[{Wang et~al.(2020)Wang, Dong, Jia, Li, Fu, Han, and
  Zhang}]{wang2020structure}
Zhiruo Wang, Haoyu Dong, Ran Jia, Jia Li, Zhiyi Fu, Shi Han, and Dongmei Zhang.
  2020.
\newblock \href {https://arxiv.org/pdf/2010.12537} {Structure-aware
  pre-training for table understanding with tree-based transformers}.
\newblock \emph{arXiv preprint arXiv:2010.12537}.

\bibitem[{Wen et~al.(2017)Wen, Vandyke, Mrk{\v{s}}i{\'c}, Ga{\v{s}}i{\'c},
  Rojas-Barahona, Su, Ultes, and Young}]{wen-etal-2017-network}
Tsung-Hsien Wen, David Vandyke, Nikola Mrk{\v{s}}i{\'c}, Milica
  Ga{\v{s}}i{\'c}, Lina~M. Rojas-Barahona, Pei-Hao Su, Stefan Ultes, and Steve
  Young. 2017.
\newblock \href {https://www.aclweb.org/anthology/E17-1042} {A network-based
  end-to-end trainable task-oriented dialogue system}.
\newblock In \emph{Proceedings of the 15th Conference of the {E}uropean Chapter
  of the Association for Computational Linguistics: Volume 1, Long Papers},
  pages 438--449, Valencia, Spain. Association for Computational Linguistics.

\bibitem[{Williams et~al.(2016)Williams, Raux, and
  Henderson}]{williams2014dialog}
Jason Williams, Antoine Raux, and Matthew Henderson. 2016.
\newblock \href
  {https://www.microsoft.com/en-us/research/publication/the-dialog-state-tracking-challenge-series-a-review/}
  {The dialog state tracking challenge series: A review}.
\newblock \emph{Dialogue \& Discourse}.

\bibitem[{Wu et~al.(2019)Wu, Madotto, Hosseini-Asl, Xiong, Socher, and
  Fung}]{wu-etal-2019-transferable}
Chien-Sheng Wu, Andrea Madotto, Ehsan Hosseini-Asl, Caiming Xiong, Richard
  Socher, and Pascale Fung. 2019.
\newblock \href {https://doi.org/10.18653/v1/P19-1078} {Transferable
  multi-domain state generator for task-oriented dialogue systems}.
\newblock In \emph{Proceedings of the 57th Annual Meeting of the Association
  for Computational Linguistics}, pages 808--819, Florence, Italy. Association
  for Computational Linguistics.

\bibitem[{Wu et~al.(2020)Wu, Zou, Jiang, and Aw}]{wu-etal-2020-gcdst}
Peng Wu, Bowei Zou, Ridong Jiang, and AiTi Aw. 2020.
\newblock \href {https://doi.org/10.18653/v1/2020.findings-emnlp.95} {{GCDST}:
  A graph-based and copy-augmented multi-domain dialogue state tracking}.
\newblock In \emph{Findings of the Association for Computational Linguistics:
  EMNLP 2020}, pages 1063--1073, Online. Association for Computational
  Linguistics.

\bibitem[{Xu et~al.(2020)Xu, Zhang, Mao, Wang, Xie, and
  Zhang}]{xu-etal-2020-curriculum}
Benfeng Xu, Licheng Zhang, Zhendong Mao, Quan Wang, Hongtao Xie, and Yongdong
  Zhang. 2020.
\newblock \href {https://doi.org/10.18653/v1/2020.acl-main.542} {Curriculum
  learning for natural language understanding}.
\newblock In \emph{Proceedings of the 58th Annual Meeting of the Association
  for Computational Linguistics}, pages 6095--6104, Online. Association for
  Computational Linguistics.

\bibitem[{Yin et~al.(2020)Yin, Shang, Jiang, Chen, and Liu}]{yin2020dialog}
Yichun Yin, Lifeng Shang, Xin Jiang, Xiao Chen, and Qun Liu. 2020.
\newblock \href {https://arxiv.org/abs/1908.07795} {Dialog state tracking with
  reinforced data augmentation}.
\newblock In \emph{Proceedings of the AAAI Conference on Artificial
  Intelligence}, volume~34, pages 9474--9481.

\bibitem[{Yoo et~al.(2020)Yoo, Lee, Dernoncourt, Bui, Chang, and
  Lee}]{yoo-etal-2020-variational}
Kang~Min Yoo, Hanbit Lee, Franck Dernoncourt, Trung Bui, Walter Chang, and
  Sang-goo Lee. 2020.
\newblock \href {https://doi.org/10.18653/v1/2020.emnlp-main.274} {Variational
  hierarchical dialog autoencoder for dialog state tracking data augmentation}.
\newblock In \emph{Proceedings of the 2020 Conference on Empirical Methods in
  Natural Language Processing (EMNLP)}, pages 3406--3425, Online. Association
  for Computational Linguistics.

\bibitem[{Yu et~al.(2021)Yu, Wu, Lin, Wang, Tan, Yang, Radev, Socher, and
  Xiong}]{yu2020grappa}
Tao Yu, Chien-Sheng Wu, Xi~Victoria Lin, Bailin Wang, Yi~Chern Tan, Xinyi Yang,
  Dragomir Radev, Richard Socher, and Caiming Xiong. 2021.
\newblock \href {https://openreview.net/forum?id=kyaIeYj4zZ} {Grappa:
  Grammar-augmented pre-training for table semantic parsing}.
\newblock \emph{International Conference on Learning Representations}.

\bibitem[{Zhang et~al.(2019)Zhang, Hashimoto, Wu, Wan, Yu, Socher, and
  Xiong}]{zhang2019find}
Jian-Guo Zhang, Kazuma Hashimoto, Chien-Sheng Wu, Yao Wan, Philip~S Yu, Richard
  Socher, and Caiming Xiong. 2019.
\newblock \href {https://arxiv.org/abs/1910.03544} {Find or classify? dual
  strategy for slot-value predictions on multi-domain dialog state tracking}.
\newblock \emph{arXiv preprint arXiv:1910.03544}.

\bibitem[{Zhao et~al.(2021)Zhao, Wang, and Huang}]{zhao2020automatic}
Yangyang Zhao, Zhenyu Wang, and Zhenhua Huang. 2021.
\newblock \href {https://arxiv.org/pdf/2012.14072} {Automatic curriculum
  learning with over-repetition penalty for dialogue policy learning}.
\newblock \emph{AAAI}.

\bibitem[{Zhong et~al.(2018)Zhong, Xiong, and Socher}]{zhong-etal-2018-global}
Victor Zhong, Caiming Xiong, and Richard Socher. 2018.
\newblock \href {https://doi.org/10.18653/v1/P18-1135} {Global-locally
  self-attentive encoder for dialogue state tracking}.
\newblock In \emph{Proceedings of the 56th Annual Meeting of the Association
  for Computational Linguistics (Volume 1: Long Papers)}, pages 1458--1467,
  Melbourne, Australia. Association for Computational Linguistics.

\bibitem[{Zhou et~al.(2020)Zhou, Yang, Wong, Wan, and
  Chao}]{zhou-etal-2020-uncertainty}
Yikai Zhou, Baosong Yang, Derek~F. Wong, Yu~Wan, and Lidia~S. Chao. 2020.
\newblock \href {https://doi.org/10.18653/v1/2020.acl-main.620}
  {Uncertainty-aware curriculum learning for neural machine translation}.
\newblock In \emph{Proceedings of the 58th Annual Meeting of the Association
  for Computational Linguistics}, pages 6934--6944, Online. Association for
  Computational Linguistics.

\bibitem[{Zhu et~al.(2020)Zhu, Li, Chen, and Yu}]{zhu-etal-2020-efficient}
Su~Zhu, Jieyu Li, Lu~Chen, and Kai Yu. 2020.
\newblock \href {https://doi.org/10.18653/v1/2020.findings-emnlp.68} {Efficient
  context and schema fusion networks for multi-domain dialogue state tracking}.
\newblock In \emph{Findings of the Association for Computational Linguistics:
  EMNLP 2020}, pages 766--781, Online. Association for Computational
  Linguistics.

\end{thebibliography}

\end{document}